\newcommand{\CLASSINPUTinnersidemargin}{0.67in} 
\newcommand{\CLASSINPUToutersidemargin}{0.67in} 
\newcommand{\CLASSINPUTtoptextmargin}{0.79in} 
\newcommand{\CLASSINPUTbottomtextmargin}{0.6in}

\documentclass[letterpaper, 10 pt, journal, twoside]{IEEEtran}
\IEEEoverridecommandlockouts

\usepackage{amsmath, amssymb, amsfonts, amsthm, mathtools, bm}
\usepackage{graphicx}
\usepackage{tabularx, threeparttable}
\usepackage{multirow, multicol}
\ifCLASSOPTIONcompsoc
\usepackage[caption=false,font=normalsize,labelfont=sf,textfont=sf]{subfig}
\else
\usepackage[caption=false,font=footnotesize]{subfig}
\fi

\usepackage[linesnumbered, lined, boxed, ruled]{algorithm2e}

\usepackage{cite}
\usepackage{textcomp}
\usepackage[dvipsnames]{xcolor}
\usepackage{url}
\usepackage{lipsum}

\newcommand{\prl}[1]{\mathopen{}\left(#1\right)\mathclose{}}

\newcommand{\crl}[1]{\mathopen{}\left\{#1\right\}\mathclose{}}

\DeclareMathOperator{\vech}{vech}

\newtheorem{theorem}{Theorem}

\newtheorem{corollary}{Corollary}[theorem]

\newtheorem{assumption}{Assumption}

\newtheorem*{proposition*}{Proposition}
\newtheorem*{theorem*}{Theorem}
\newtheorem*{corollary*}{Corollary}

\usepackage{scalerel,stackengine}
\stackMath
\newcommand\reallywidehat[1]{
\savestack{\tmpbox}{\stretchto{
  \scaleto{
    \scalerel*[\widthof{\ensuremath{#1}}]{\kern.1pt\mathchar"0362\kern.1pt}%
    {\rule{0ex}{\textheight}}
  }{\textheight}
}{2.4ex}}
\stackon[-6.9pt]{#1}{\tmpbox}
}

\def\BibTeX{{\rm B\kern-.05em{\sc i\kern-.025em b}\kern-.08em
    T\kern-.1667em\lower.7ex\hbox{E}\kern-.125emX}}

\begin{document}

\copyright 2020 IEEE. Personal use of this material is permitted.
Permission from IEEE must be obtained for all other uses, in any current or future media, including reprinting/republishing this material for advertising or promotional purposes, creating new collective works, for resale or redistribution to servers or lists, or reuse of any copyrighted component of this work in other works.

\title{Belief Space Planning for Mobile Robots with Range Sensors using iLQG}

\author{Ke~Sun and~Vijay~Kumar
\thanks{Manuscript received: Oct, 14, 2020; Revised Jan, 1, 2021; Accepted Feb, 6, 2021.}
\thanks{This paper was recommended for publication by Editor Nancy Amato upon evaluation of the Associate Editor and Reviewers' comments.
This work was supported by ARL Grant DCIST CRA W911NF-17-2-0181,
                           NSF Grant CNS-1521617,
                           ARO Grant W911NF-13-1-0350,
                           ONR Grants N00014-20-1-2822,
                           ONR grant N00014-20-S-B001,
                           and Qualcomm Research.}
\thanks{Ke Sun and Vijay Kumar are with GRASP Lab,
        University of Pennsylvania, Philadelphia, PA 19104, USA,
        {\tt \footnotesize\{sunke, kumar\}@seas.upenn.edu}}
\thanks{Digital Object Identifier (DOI): see top of this page.}
}


\maketitle

\begin{abstract}
  In this work, we use iterative Linear Quadratic Gaussian (iLQG) to plan motions for a mobile robot with range sensors in belief space.
  We address two limitations that prevent applications of iLQG to the considered robotic system.
  First, iLQG assumes a differentiable measurement model, which is not true for range sensors.
  We show that iLQG only requires the differentiability of the belief dynamics.
  We propose to use a derivative-free filter to approximate the belief dynamics, which does not require explicit differentiability of the measurement model.
  Second, informative measurements from a range sensor are sparse.
  Uninformative measurements produce trivial gradient information, which prevent iLQG optimization from converging to a local minimum.
  We densify the informative measurements by introducing additional parameters in the measurement model.
  The parameters are iteratively updated in the optimization to ensure convergence to the true measurement model of a range sensor.
  We show the effectiveness of the proposed modifications through an ablation study.
  We also apply the proposed method in simulations of large scale real world environments, which show superior performance comparing to the state-of-the-art methods that either assume the separation principle or maximum likelihood measurements.
\end{abstract}

\begin{IEEEkeywords}
Motion and Path Planning, Sensor-based Control
\end{IEEEkeywords}

\section{Introduction}
\label{sec: introduction}
\IEEEPARstart{M}{otion} planning is essential for autonomy of a robot in completing a task.
In practice, robotic systems are subject to unmodeled dynamics or environment, which is often compensated by introducing random processes, \textit{i.e.} noise, to the system model.
The noise needs to be explicitly considered in the motion planning in order to complete a task reliably.
In this work, we refer to such motion planning problems as stochastic motion planning.
More specifically, we consider the problem of navigating a car-like robot with a range sensor.

A principled way to solve stochastic motion planning is to model the problem as a Partially Observed Markov Decision Process (POMDP)~\cite{Kaelbling-AI-1998}.
Methods following this approach are often known as belief space planning in the robotics community.
van den Berg \textit{et al.}~\cite{vandenBerg-IJRR-2012} extend iterative Linear Quadratic Regulator (iLQR)~\cite{Li-ICINCO-2004} to belief dynamics, and propose (belief space) iterative Linear Quadratic Gaussian (iLQG), which solves a local optimal control policy for a continuous POMDP.
Superior performance is shown in~\cite{vandenBerg-IJRR-2012} for light-dark domain problems, where measurements have small or large noise in light or dark regions respectively.
However, few work has reported successful applications of iLQG for more common sensors, such as range sensors considered in this work.

We summarize the reasons for the lack of applications of iLQG to range sensors as follows.
First, iLQG requires the underlying system to have differentiable motion and measurement models.
However, because of the discontinuity in the environment, the range sensor model is often nondifferentiable.
Second, informative measurements, \textit{i.e.} measurements that are effective in reducing localization uncertainty, from range sensors are sparse.
To elaborate, informative measurements can only be obtained if obstacles are within the maximum sensing range of the beams.
If the robot is sufficiently far from the obstacles with all measurements saturated at the maximum range, it cannot be known where to move to collect informative measurements by just locally perturbing the robot state.
In terms of optimization, it means a majority of states are saddle points, providing trivial gradient information and preventing iLQG from converging to a local minimum.

\textbf{Contributions:} In this work, we apply iLQG to solve the stochastic motion planning problem for car-like robots with range sensors by addressing the previously mentioned two limitations.
Our contributions are summarized as follows:

First, the application of iLQG only requires the belief dynamics to be differentiable.
We show that the differentiability of the true discrete-time belief dynamics, modeled by a Bayes filter, is independent of the underlying motion or measurement models.
The explicit requirement for differentiability in~\cite{vandenBerg-IJRR-2012} is because of the use of an Extended Kalman Filter (EKF) in approximating the belief dynamics.
To overcome this issue, we propose to use a derivative-free filter based on an Unscented Kalman Filter (UKF) for belief dynamics modeling.

Second, the discontinuity of the noise standard deviation of a range sensor model can be understood as the cause of sparse informative measurements.
The noise standard deviation is nominal when the measurement is within the sensing range, which is otherwise infinity.
We propose to approximate the step function with a sigmoid function, which is equivalent to introducing an artificial gradient into the optimization, therefore, densifying the informative measurements.
By properly scheduling the sigmoid parameters in an outer loop of iLQG, the modified measurement model eventually converges to the true model, ensuring that the resulting control policy is designed for the original system.

In the experiments, we demonstrate the effectiveness of the proposed modifications through an ablation study.
The proposed method is also applied to large scale real world environments mapped with real Lidar data.
We show the superior performance of the proposed method by comparing it with state-of-the-art methods that either assumes the separation principle or maximum likelihood measurements.

\section{Related Works}
\label{sec: related works}
Most robotic applications~\cite{Thrun-JFR-2006, Salavasidis-JFR-2019, Mohta-JFR-2018} that involve perception action loops invoke the separation principle~\cite[Ch.4]{Bertsekas-Athena-1995} to simplify the development of planning/control algorithms, which is hard to justify for nonlinear systems.
Relying on the separation principle implies ignoring the stochasticity in state estimation.
Albeit the advancement of estimation algorithms with various sensors, the estimation may not be sufficiently accurate due to environmental conditions or theoretical limitations.
The mismatch between the estimated and the true state can lead to undesired consequences such as collisions.
Recent works on perception-aware planning~\cite{Falanga-IROS-2018, Spasojevic-ICRA-2020} attempt to address this issue by introducing perception-related costs into the objective function of planning.
The resulting trajectories demonstrate active localization behavior.
However, the relationship between the heuristically defined perception costs and the actual estimation uncertainty remains unclear.

As mentioned in Sec.~\ref{sec: introduction}, belief space planning is a principled way of solving stochastic motion planning problems, which directly plans for the belief instead of the state of a system.
The major difficulty is that all future measurement sequences have to be rollout in order to determine an optimal policy.
A large body of literature~\cite{Prentice-IJRR-2009, Platt-RSS-2010, Kopitkov-IJRR-2019} simplifies the problem with maximum likelihood measurement assumption, where only the maximum likelihood measurement sequence is evaluated.
Du Toit~\cite{DuToit-TRO-2012} shows that the assumption introduces artificial information, which makes the resulting control policy less robust if the actual future measurements differ from the ones with maximum likelihood.

Related to the modifications introduced in this work, van den Berg~\cite{vandenBerg-IJRR-2012} briefly mentions that UKFs could replace EKFs to approximate belief dynamics.
Nishimurai~\cite{Nishimurai-WAFR-2018} also applies a UKF as belief dynamics approximation in an active multi-target tracking problem with (unlimited) range measurements.
In both works, the application of UKFs are not intended for nondifferentiable system models.
Bachrach~\cite{Bachrach-IJRR-2012} shares the same spirit with this work, where a UKF is applied in the belief roadmap (BRM) framework~\cite{Prentice-IJRR-2009} for nondifferentiable measurement models of RGB-D cameras.
One technical flaw cannot be avoided in~\cite{Bachrach-IJRR-2012}.
With the application of UKF in BRM, the computation of ``transfer functions'' of covariance requires the prior uncertainty, which is unavailable while constructing the roadmap.

A few works~\cite{Patil-ICRA-2014, Indelman-IJRR-2015, Chaves-IROS-2015} consider randomness in measurement acquisition, which is similar to the sparse informative measurements problem.
Patil \textit{et al.}~\cite{Patil-ICRA-2014} introduces signed distance field to model sensing regions.
The use of signed distance field implicitly assumes the direction of movement for obtaining measurements is unique and known \textit{a priori}, which is not true for range sensors.
Indelman \textit{et al.}~\cite{Indelman-IJRR-2015} and Chaves \textit{et al.}~\cite{Chaves-IROS-2015} introduce Bernoulli random variables to model the randomness in measurement acquisition.
In~\cite{Indelman-IJRR-2015}, the binary random variable is computed with the expected state and is fixed during the optimization.
In~\cite{Chaves-IROS-2015}, the random variable is assumed to be independent of the state.
The issue of sparse informative measurements cannot be addressed by either method because of the assumed independence between the state and measurement acquisition during the optimization.

\section{Preliminaries}
\label{sec: preliminary}
\begin{algorithm}[t]
  \DontPrintSemicolon
  \SetCommentSty{emph}
  \SetKwProg{Fn}{Function}{}{end}
  \KwIn{initial policy $\{\bar{\bm{b}}_0, \pi_0, \dots, \bar{\bm{b}}_l\}$.}
  \KwOut{local optimal policy $\{\pi^*_0, \pi^*_1, \dots, \pi^*_{l-1}\}$.}

  Initialize $\crl{\pi^*_0, \dots, \pi^*_{l-1}}$ with $\crl{\pi_0, \dots, \pi_{l-1}}$.\;

  \While{policy has not converged} {
    Generate the nominal trajectory $\crl{\bar{\bm{b}}_0, \bar{\bm{u}}_0, \dots, \bar{\bm{b}}_l}$ with $\bar{\bm{u}}_{t} = \pi^*_t(\bar{\bm{b}}_t)$.\;
    Approximate the final step optimal value function $V^*_l$ with a second-order Taylor expansion of $c_l$ in~\eqref{eq: objective function} around $\bar{\bm{b}}_l$.\;
    \For{$t$ from $l-1$ to $0$}{
      Approximating $\Phi$ in~\eqref{eq: EKF belief dynamics} with a first-order Taylor expansion around $\bar{\bm{b}}_t$ and $\bar{\bm{u}}_t$ to produce a linear belief dynamics $\Phi'$.\;
      Approximating $c_t$ in~\eqref{eq: objective function} with a second-order Taylor expansion around $\bar{\bm{b}}_t$ and $\bar{\bm{u}}_t$ to produce a quadratic stage cost $c'_t$.\;
      Solve for $\pi^*_t$ and $V^*_{t}$ through minimizing $c'_t + \mathbb{E}\{V^*_{t+1}(\Phi')\}$.\;
    }
  }

  \caption{iLQG}
  \label{alg: belief space iLQG}
\end{algorithm}

In this section, we briefly review iLQG in~\cite{vandenBerg-IJRR-2012}.
Consider a discrete-time system with motion and measurement models in the form of,
\begin{equation}
  \label{eq: general discrete time system motion and measurement models}
  \begin{aligned}
    \bm{x}_{t+1} &= f\prl{\bm{x}_t, \bm{u}_t, \bm{m}_t},\; &\bm{m}_t\sim\mathcal{N}(\bm{0}, \Sigma_{\bm{m}}), \\
    \bm{z}_t     &= h\prl{\bm{x}_t, \bm{n}_t},\;           &\bm{n}_t\sim\mathcal{N}(\bm{0}, \Sigma_{\bm{n}}).
  \end{aligned}
\end{equation}
where $\bm{x}$, $\bm{u}$, $\bm{z}$ are the state, control, and measurement.
$\bm{m}$ and $\bm{n}$ are the Gaussian motion and measurement noises.

If both motion and measurement models are differentiable, an EKF can be applied to approximate the belief dynamics.
In EKFs, belief is approximated with Gaussian distributions, $\mathcal{N}\prl{\hat{\bm{x}}, \Sigma}$, where $\hat{\bm{x}}$ is the estimated mean of the state, $\Sigma$ is the covariance.
With the latest controls and measurements, the belief is updated as,
\begin{equation}
  \label{eq: extended kalman filter}
  \begin{aligned}
    \hat{\bm{x}}_{t+1} &= f(\hat{\bm{x}}_t, \bm{u}_t, \bm{0}) +
                         K_{t+1} \prl{\bm{z}_{t+1} - h(f(\hat{\bm{x}}_t, \bm{u}_t, \bm{0}), \bm{0})}, \\
    \sqrt{\Sigma_{t+1}} &= \sqrt{\Gamma_t - K_{t+1} H_{t+1} \Gamma_t}.
  \end{aligned}
\end{equation}
where,
\begin{equation*}
  \begin{gathered}
    \begin{aligned}
      &\Gamma_t = A_t \Sigma_t A_t^\top + M_t M_t^\top, \\
      &K_{t+1}  = \Gamma_t H_{t+1}^\top (H_{t+1} \Gamma_t H_{t+1}^\top + N_{t+1} N_{t+1}^\top)^{-1},
    \end{aligned} \\
    \begin{aligned}
      &A_t     = \frac{\partial f}{\partial \bm{x}} \prl{\hat{\bm{x}}_t, \bm{u}_t, \bm{0}}, &
      &M_t     = \frac{\partial f}{\partial \bm{m}} \prl{\hat{\bm{x}}_t, \bm{u}_t, \bm{0}}, \\
      &H_{t+1} = \frac{\partial h}{\partial \bm{x}} \prl{f(\hat{\bm{x}}_t, \bm{u}_t, \bm{0}), \bm{0}}, &
      &N_{t+1} = \frac{\partial h}{\partial \bm{n}} \prl{f(\hat{\bm{x}}_t, \bm{u}_t, \bm{0}), \bm{0}}.
    \end{aligned}
  \end{gathered}
\end{equation*}
Instead of $\Sigma$, its square root, $\sqrt{\Sigma}$ (the Cholesky factorization of $\Sigma$), is used for numerical stability.
We may also take advantage of the sparsity of $\sqrt{\Sigma}$, and represent the belief in a vector form, $\bm{b} = \prl{\hat{\bm{x}}^\top, \vech(\sqrt{\Sigma})^\top}^\top$, where $\vech(\sqrt{\Sigma})$ stacks the lower (or equivalently upper) triangular entries of $\sqrt{\Sigma}$ as a vector.
With the new belief representation, the belief dynamics in~\eqref{eq: extended kalman filter} is rewritten as,
\begin{equation}
  \label{eq: EKF belief dynamics}
    \bm{b}_{t+1} =
    \begin{pmatrix}
      f(\hat{\bm{x}}_t, \bm{u_t}, \bm{0}) + \sqrt{K_{t+1} H_{t+1} \Gamma_t}\bm{w}_{t+1} \\
      \vech\prl{\sqrt{\Gamma_t - K_{t+1} H_{t+1} \Gamma_t}}
    \end{pmatrix},
\end{equation}
where $\bm{w}_{t+1} \sim \mathcal{N}(\bm{0}, I)$ is the normalized innovation noise.

The objective is to find the control policy $\bm{u}_t = \pi_t(\bm{b}_t)$ that minimizes the following summation of cost functions over a finite horizon,
\begin{equation}
  \label{eq: objective function}
  V_0(\bm{b}_0) = \mathbb{E} \crl{\sum_{t=0}^{l-1} c_t(\bm{b}_t, \bm{u}_t) + c_l(\bm{b}_l)},
\end{equation}
where $c_t(\cdot)$'s and $c_l(\cdot)$ are assumed to be second-order differentiable with positive-(semi)definite Hessians.

The key observation in~\cite{vandenBerg-IJRR-2012} is that an EKF has the form,
\begin{equation}
  \label{eq: general belief dynamics}
  \bm{b}_{t+1} = \Phi(\bm{b}_t, \bm{u}_t, \bm{z}_{t+1}),
\end{equation}
sharing the same form as the motion model in~\eqref{eq: general discrete time system motion and measurement models}.
Specially, $\bm{b}$ is the (belief) state, $\bm{u}$ is the control input, and $\bm{z}$ plays the role of motion noise.
Therefore, one can apply iLQR~\cite{Li-ICINCO-2004} on~\eqref{eq: general belief dynamics} to obtain a local optimal feedback policy of belief.

The general framework of iLQG is summarized in Alg.~\ref{alg: belief space iLQG}.
Given an initial control policy, the original nonlinear optimization problem is converted to an LQG by approximating the belief dynamics and the cost functions through first and second order Taylor expansions respectively.
The control policy for the LQG can be solved recursively through dynamic programming.
With the new control policy, the nominal trajectory is updated which is then used to update the approximations for the belief dynamics and costs for the next iteration.
We omit steps like line search or trust region methods in Alg.~\ref{alg: belief space iLQG} for brevity.
Details of the iLQG steps can be found in~\cite{vandenBerg-IJRR-2012}.

\section{Problem Formulation}
\label{sec: problem definition}
In this work, we consider the problem of navigating a car-like mobile robot with range sensors in a known environment.
We assume the mobile robot is controlled with velocity command. The discrete-time dynamics is,
\begin{equation}
  \label{eq: car-like mobile robot model}
  \begin{gathered}
    \bm{\xi} = \begin{pmatrix}1 & 0 \\ 0 & 0 \\ 0 & 1\end{pmatrix} (\bm{u}_t + \bm{m}_t),\;
    \bm{m}_t \sim \mathcal{N}\prl{\bm{0}, \begin{pmatrix}\sigma_v & 0 \\ 0 & \sigma_\omega\end{pmatrix}},\\
    \bm{x}_{t+1} = f(\bm{x}_t, \bm{u}_t, \bm{m}_t)
                 = \bm{x}_t \circ \exp \prl{\bm{\xi} \cdot \tau}. \\
  \end{gathered}
\end{equation}
In~\eqref{eq: car-like mobile robot model}, $\bm{x}\in SE(2)$ is the state of the mobile robot.
$\bm{u}=\prl{v, \omega}^\top\in\mathbb{R}^2$ is the control input, consisting of linear velocity, $v$, and angular velocity, $\omega$.
$\bm{m}$ is the $i.i.d.$ motion noise.
$\bm{\xi}$ is the 2-D twist.
$\tau$ is the duration of each discrete time step.
Finally, $\circ$ and $\exp(\cdot)$ denote the composition and exponential operations on $SE(2)$.
Note that, for brevity in here and the following, we omit $\hat{\cdot}$ or $\check{\cdot}$ operations that convert elements in Lie algebra between matrix and vector representations.

The mobile robot is equipped with a range sensor (\textit{e.g.} Lidars or RGB-D Cameras) for localization.
The range sensor is modeled as follows,
\begin{equation}
  \label{eq: range sensor model}
  \begin{aligned}
    \bm{n}_t &= \prl{n^1_t, n^2_t, \dots, n^m_t}^\top \sim \mathcal{N}\prl{\bm{0}, I}, \\
    z^i_t  &= r(\bm{x}_t, \bm{o}, \theta^i) + N\prl{r(\bm{x}_t, \bm{o}, \theta^i)} n^i_t, \; i = 1, \dots, m, \\
    N(r) &=
    \begin{cases}
      \sigma_n & \text{if } r < r_m, \\
      \infty   & \text{otherwise},
    \end{cases} \\
    \bm{z}_t &= h(\bm{x}_t, \bm{n}_t) = \prl{z^1_t, z^2_t, \dots, z^m_t}^\top.
  \end{aligned}
\end{equation}
In~\eqref{eq: range sensor model}, $\bm{z}\in\mathbb{R}^m$ consists of range measurements from $m$ beams.
$\bm{n}$ is the \textit{i.i.d} measurement noise.
$\bm{o}$ is the known map in the form of occupancy grid.
$r(\bm{x}, \bm{o}, \theta)$ models the ray casting process of a beam oriented at $\theta$ in the frame $\bm{x}$.
The noise standard deviation of a range measurement is scaled by $N(r)$.
The noise has standard deviation $\sigma_n$ if the predicted range $r$ is within the maximum sensing range $r_m$.
Otherwise, the noise standard deviation is infinite.
Note that the measurement model in~\eqref{eq: range sensor model} does not exactly match the behavior of a range sensor in practice when $r\geq r_m$.
In the cases that $r \geq r_m$, range measurements often saturate at $r_m$.
In this work, the measurement model is only used to construct belief dynamics.
Setting the noise standard deviation to infinity means that measurements with $r \geq r_m$ have no effect in belief update.
The same assumption is often applied in practical range sensor based estimation algorithms~\cite{Pomerleau-AR-2013}.

With the measurement model in~\eqref{eq: range sensor model},  an EKF, as used in Sec.~\ref{sec: preliminary}, is no longer suitable to model the belief dynamics.
Since the environment modeled by $\bm{o}$ is often discontinuous, the ray casting function $r(\cdot)$ in~\eqref{eq: range sensor model} is nondifferentiable.
For now, we simply leave the belief dynamics in its general form~\eqref{eq: general belief dynamics}.
The actual function $\Phi(\cdot)$ will be specified in Sec.~\ref{sec: approach}.

As in Sec.~\ref{sec: preliminary}, we assume the motion planning task can be modeled by an objective function formulated as~\eqref{eq: objective function}, where the stage costs $c_t(\cdot)$'s and terminal cost $c_l(\cdot)$ are second order differentiable with positive-(semi)definite Hessians.
A control policy is then to be solved by optimizing the objective.

\section{Approach}
\label{sec: approach}
In this section, we address the issues that prevent the application of iLQG to system defined in Sec.~\ref{sec: problem definition}, namely the nondifferentiable measurement model and sparse informative measurements.
At the end of this section, we also introduce the cost functions used in this work to model the motion planning task in an obstacled environment.

\subsection{UKF Belief Dynamics}
\label{subsec: UKF belief dynamics}
The key insight is that the application of iLQG only requires differentiability of the belief dynamics in~\eqref{eq: general belief dynamics}.
The requirement of explicit differentiability of the motion and measurement models in~\cite{vandenBerg-IJRR-2012} is because of the usage of an EKF to approximate belief dynamics.
In Appendix, we show that the true belief dynamics, modeled by a Bayes filter, is differentiable with weak assumptions.
The assumptions are trivially satisfied in~\eqref{eq: general discrete time system motion and measurement models} since system noises follow Gaussian distributions.
In this work, we use a derivative free filter, a UKF, to model the belief dynamics, which avoids explicitly differentiating the motion or the measurement models.

In an (on-manifold) UKF~\cite{Brossard-IROS-2017}, belief, $\mathcal{N}\prl{\bm{x}, \Sigma}$, is approximated with $2n+1$ sigma points with $n$ for the state dimension.
\begin{equation}
  \label{eq: sigma points}
  \begin{aligned}
    &\bm{s}^i = \bm{x} \circ \exp\prl{\sqrt{(n+\lambda)\Sigma}^{(i)}},\\
    &\bm{s}^{i+n} = \bm{x} \circ \exp\prl{-\sqrt{(n+\lambda)\Sigma}^{(i)}},\\
    &\bm{s}^0 = \bm{x},\;
    i = 1, 2, \dots, n.
  \end{aligned}
\end{equation}
In~\eqref{eq: sigma points}, $\lambda=\alpha^2(n+\kappa)-n$, where $\alpha$ and $\kappa$ control the spread of the sigma points.
The notation $A^{(i)}$ refers to the $i$\textsuperscript{th} column in matrix $A$.
The weights for the sigma points are,
\begin{equation}
  \label{eq: sigma point weights}
  \begin{aligned}
    &w_m^0 = \frac{\lambda}{n+\lambda},\quad
    w_m^i = \frac{\lambda}{n+\lambda} + (3-\alpha^2), \\
    &w_c^0 = w_c^i = \frac{1}{2(n+\lambda)}, \; i = 1, 2, \dots, 2n.
  \end{aligned}
\end{equation}
$w_m$'s and $w_c$'s are used to recover the first and second moments of the Gaussian distribution respectively.
We define $\mathcal{S} = s(\bm{x}, \Sigma)$ with $\mathcal{S} = \crl{\prl{\bm{s}^i, w_m^i, w_c^i}, i=0, 1, \dots, 2n}$, to represent the process of generating sigma points,~\eqref{eq: sigma points}, and the weights,~\eqref{eq: sigma point weights}, from a Gaussian distribution, $\mathcal{N}\prl{\bm{x}, \Sigma}$.

At the prediction step, sigma points generated with $\mathcal{N}\prl{\hat{\bm{x}}_t, \Sigma_t}$ are propagated through the motion model to obtain the prior belief $\mathcal{N}\prl{\bar{\bm{x}}_{t+1}, \bar{\Sigma}_{t+1}}$,
\begin{equation}
  \label{eq: UKF prediction step}
  \begin{gathered}
    \mathcal{S}_t = s(\hat{\bm{x}}_t, \Sigma_t), \quad
    \bm{\xi}_t^i = \log\prl{f^{-1}(\bm{s}^0_t, \bm{u}_t, \bm{0}) \circ f(\bm{s}^i_t, \bm{u}_t, \bm{0})},\\
    \begin{aligned}
      \bar{\bm{x}}_{t+1} &= f(\bm{s}^0_t, \bm{u}_t, \bm{0}) \circ
                           \exp\prl{\sum_{i=0}^{2n} w_m^i \bm{\xi}_t^i},\\
      \bar{\Sigma}_{t+1} &= \sum_{i=0}^{2n} w_c^i \bm{\xi}_t^i \bm{\xi}_t^{i\top} + M_t \Sigma_{\bm{m}} M_t^\top,
    \end{aligned}
  \end{gathered}
\end{equation}
with $M_t = \partial f(\hat{\bm{x}}_t, \bm{u}_t, \bm{0}) / \partial \bm{m}_t$.

At the update step, the posterior belief is obtained based on the difference between the actual measurement and the measurements predicted with the sigma points,
\begin{equation}
  \label{eq: UKF update step}
  \begin{gathered}
    \begin{aligned}
      &\bar{\mathcal{S}}_{t+1} = s(\bar{\bm{x}}_{t+1}, \bar{\Sigma}_{t+1}), &\quad
      &\bm{\xi}^i_{t+1} = \bar{\bm{x}}_{t+1}^{-1} \circ \bar{\bm{s}}^i_{t+1},\\
      &\bar{\bm{z}}_{t+1} = \sum_{i=0}^{2n} w_m^i h\prl{\bar{\bm{s}}^i_{t+1}, \bm{0}}, &\quad
      &\bm{\delta}^i_{t+1} = h\prl{\bar{\bm{s}}^i_{t+1}, \bm{0}} - \bar{\bm{z}}_{t+1},
    \end{aligned} \\
    \begin{aligned}
      &V_{t+1} = \sum_{i=0}^{2n} w_c^i \bm{\delta}^i_{t+1} \bm{\delta}_{t+1}^{i\top} + N_{t+1} \Sigma_{\bm{n}} N_{t+1}^\top, \\
      &P_{t+1} = \sum_{i=0}^{2n} w_c^i \bm{\xi}^i_{t+1} \bm{\delta}_{t+1}^{i\top}, \quad
      K_{t+1} = P_{t+1} V_{t+1}^{-1}, \\
      &\hat{\bm{x}}_{t+1} = \bar{\bm{x}}_{t+1} \circ \exp\prl{K_{t+1} \prl{\bm{z}_{t+1} - \bar{\bm{z}}_{t+1}}}, \\
      &\Sigma_{t+1} = \bar{\Sigma}_{t+1} - K_{t+1} V_{t+1} K_{t+1},
    \end{aligned}
  \end{gathered}
\end{equation}
with $N_{t+1} = \partial h(\bar{\bm{x}}_{t+1}, \bm{0}) / \partial \bm{n}_{t+1}$.

By representing the belief $\mathcal{N}\prl{\hat{\bm{x}}, \Sigma}$ in the vector form $\prl{\hat{\bm{x}}^\top, \vech\prl{\Sigma}^\top}^\top$, the belief dynamics modeled by a UKF can be rewritten compactly as,
\begin{equation}
  \label{eq: UKF belief dynamics}
  \bm{b}_{t+1} =
  \begin{pmatrix}
    \bar{\bm{x}}_{t+1} \circ \exp\prl{\sqrt{K_{t+1} V_{t+1} K^\top_{t+1}} \bm{w}_{t+1}} \\
    \vech\prl{\bar{\Sigma}_{t+1} - K_{t+1} V_{t+1} K^\top_{t+1}}
  \end{pmatrix}.
\end{equation}
Note that~\eqref{eq: UKF belief dynamics} is in the same form of~\eqref{eq: EKF belief dynamics}.
Therefore, the iLQG as in Alg.~\ref{alg: belief space iLQG} can be applied seamlessly.

\subsection{Densify Informative Measurements}
\label{subsec: densify informative measurements}
As discussed in Sec.~\ref{sec: introduction}, applying iLQG naively with~\eqref{eq: range sensor model} may not produce the desired control policy because of the sparsity of informative measurements.
The issue can be better understood through the noise standard deviation $N(\cdot)$ in~\eqref{eq: range sensor model}.
$N(\cdot)$ is a step function (assuming infinity is replaced by a large enough finite value), the gradient of which is zero almost everywhere except at $r_m$.
The lack of nonzero gradient of $N(\cdot)$ makes it hard for iLQG to locally determine controls that lead to informative measurements.

In this work, we address this issue by approximating $N(\cdot)$ with a sigmoid function shown as the following,
\begin{equation}
  \label{eq: sigmoid function for noise scale}
  N_s(r) = \frac{\mu}{1+e^{-\nu(r-r_m)}} + \sigma_n.
\end{equation}
Recall that, in~\eqref{eq: range sensor model}, $r_m$ is the maximum sensing range. $\sigma_n$ is the noise standard deviation when $r<r_m$.
In~\eqref{eq: sigmoid function for noise scale}, $\nu$ controls the gradient of $N_s(\cdot)$.
By increasing $\nu$, $N_s(\cdot)$ is closer to a step function.
$\mu$ controls the maximum value of $N_s(\cdot)$.
Ideally, $\mu$ should be kept at a large value to simulate large noise standard deviation when $r\geq r_m$.
However, consider the case of having a large $\mu$ but a small $\nu$ in~\eqref{eq: sigmoid function for noise scale}.
$N_s(r)$ would be large for all $r\geq 0$, making all measurements uninformative.
Therefore, it is important to also set $\mu$ as a parameter which gradually approaches infinity with $\nu$ during the optimization.

\begin{algorithm}[t]
  \DontPrintSemicolon
  \SetCommentSty{emph}
  \SetKwProg{Fn}{Function}{}{end}
  \KwIn{initial policy $\{\bar{\bm{b}}_0, \pi_0, \dots, \bar{\bm{b}}_l\}$.}
  \KwOut{local optimal policy $\{\pi^*_0, \pi^*_1, \dots, \pi^*_{l-1}\}$.}

  \tcp{$\mu_0$ and $\nu_0$ are predefined initial values.}
  $\mu \leftarrow \mu_0$, $\nu \leftarrow \nu_0$\;

  \tcp{$\mu_m$ and $\nu_m$ are the predefined upper bounds.}
  \While{$\mu < \mu_m$ and $\nu < \nu_m$} {
    Apply Alg.~\ref{alg: belief space iLQG} to obtain $\{\pi^*_0, \pi^*_1, \dots, \pi^*_{l-1}\}$.\;
    $\mu \leftarrow \lambda_{\mu} \mu$, $\nu \leftarrow \lambda_{\nu} \nu$, $\prl{\lambda_{\mu}, \lambda_{\nu} > 1}$.\;
    Update the belief in the nominal trajectory of the feedback policy with the new measurement model.
  }

  \caption{iLQG with modified range sensor model}
  \label{alg: belief space iLQG with modified range sensor model}
\end{algorithm}

With the modified range sensor model, iLQG can be applied. 
As shown in Alg.~\ref{alg: belief space iLQG with modified range sensor model}, an outer loop, controlling the change of $\mu$ and $\nu$, wraps around the original iLQG algorithm in Alg.~\ref{alg: belief space iLQG}.
By increasing both $\mu$ and $\nu$ monotonically, $N_s(\cdot)$ converges to $N(\cdot)$.
(Numerically, $N(\cdot)$ in~\eqref{eq: range sensor model} is not well defined because of the involvement of $\infty$. Here we assume $N(\cdot)$ can be approximated by $N_s(\cdot)$ at $\mu=\mu_m$ and $\nu = \nu_m$ with sufficient accuracy.)
Therefore, the measurement model in iLQG would eventually converge to~\eqref{eq: range sensor model}.

\subsection{Cost Functions}
\label{subsec: cost functions}
To model the motion planning task, the cost functions used in this work include three kinds of objectives, namely reaching the goal, minimizing control effort, and avoiding collisions.
More specifically, the stage cost is in the form of,
\begin{equation}
  \label{eq: stage cost}
  \begin{aligned}
  c_t\prl{\bm{b}_t, \bm{u}_t}
    =& \log^{\top}(\bm{b}^{-1}_t \circ \bm{b}_g) \cdot Q^{\bm{b}}_t \cdot \log(\bm{b}^{-1}_t \circ \bm{b}_g)
        + \bm{u}^\top_t R_t \bm{u}_t \\
     &- q_c \cdot \log\prl{\Gamma^{-1}(1) \cdot \gamma\prl{1, \frac{\sigma^2(\bm{b}_t, \bm{o})}{2}}} \\
     &+ q_d \cdot d^{-1}(\bm{b}_t, \bm{o}).
  \end{aligned}
\end{equation}

In~\eqref{eq: stage cost}, the first two terms are quadratic costs trying to reduce the distance to the goal belief state, and the control magnitude.
We approximate the belief space $\mathcal{B}$ with $SE(2)\times\mathbb{R}^6$, with $SE(2)$ and $\mathbb{R}^6$ for mean and uncertainty respectively.
Compared to approximating $\mathcal{B}$ with $\mathbb{R}^9$, which combines mean and uncertainty into a single vector space, $SE(2)\times\mathbb{R}^6$ respects the kinematic constraints of the state space, therefore, can more accurately reflect the difference between belief states.

To avoid collisions, we use the method proposed in~\cite{vandenBerg-IJRR-2012} to approximate the collision probability.
The function $\sigma(\bm{b}, \bm{o})$ is defined as $\min_{\bm{c}} \left\|\hat{\bm{t}}-\bm{c}\right\|_{\Sigma}$, the minimum normalized distance (normalized with the uncertainty $\Sigma$) between the position estimate $\hat{\bm{t}}$ and occupied cells $\bm{c}$ in $\bm{o}$.
With $\sigma(\bm{b}, \bm{o})$, the regularized gamma function, $\Gamma^{-1}(1)\gamma(\cdot)$, provides a lower bound for the probability of not colliding with obstacles.
The optimization increases the clearance probability by minimizing the negative logarithm of the regularized gamma function.

In practice, we find the collision probability cost may degenerate when the uncertainty, $\Sigma$, is small.
For the extreme case when $\Sigma = 0$, the collision probability cost is zero regardless of the robot position.
To resolve this issue, we introduce an additional cost, $d^{-1}(\bm{b}, \bm{o})$, depending on the absolute Euclidean distance between the position estimate and the occupied cells.
$d(\bm{b}, \bm{o})$ is defined as $\min_{\bm{c}} \|\hat{\bm{t}}-\bm{c}\|_2$.

The terminal cost $c_l$ is defined in a similar way as $c_t$,
The minor differences between $c_l$ and $c_t$ include the change of parameter, such as $Q^{\bm{b}}$, and the removal of control effort cost.

\section{Experiments}
\label{sec: experiments}
For all simulations, the resolution of the occupancy grid map, \textit{i.e.} cell size, is assumed to be $0.1$m.
In the motion model~\eqref{eq: car-like mobile robot model}, the timer interval is set to $\tau=0.1$s, motion noise is set to $\sigma_v=0.5$m/s, $\sigma_{\omega}=0.05$rad/s.
We assume the application of a low-cost range sensor.
The range sensor has five beams oriented at $\crl{-\pi/2, -\pi/4, 0, \pi/4, \pi/2}$ with maximum range $r_m=2$m.
The measurement noise in~\eqref{eq: range sensor model} is set to $\sigma_z=0.5$m.
Sigmoid function parameters $\mu_m$ and $\nu_m$, in Alg.~\ref{alg: belief space iLQG with modified range sensor model}, are both fixed at $1e3$ for all simulations, while $\mu_0$, $\nu_0$, $\lambda_{\mu}$, and $\lambda_{\nu}$ are reported for individual cases.
The initial trajectories are created with stable sparse RRT~\cite{Li-IJRR-2016} implemented in the open motion planning library~\cite{Sucan-RAM-2012}.
The algorithms are timed on a laptop scale computer with Intel i7-6670HQ CPU (4 cores at 2.6GHz) and 32GB RAM.

\subsection{Ablation study}
\label{subsec: ablation study}
\begin{figure}
  \centering
  \subfloat[]{\includegraphics[width=0.25\columnwidth]{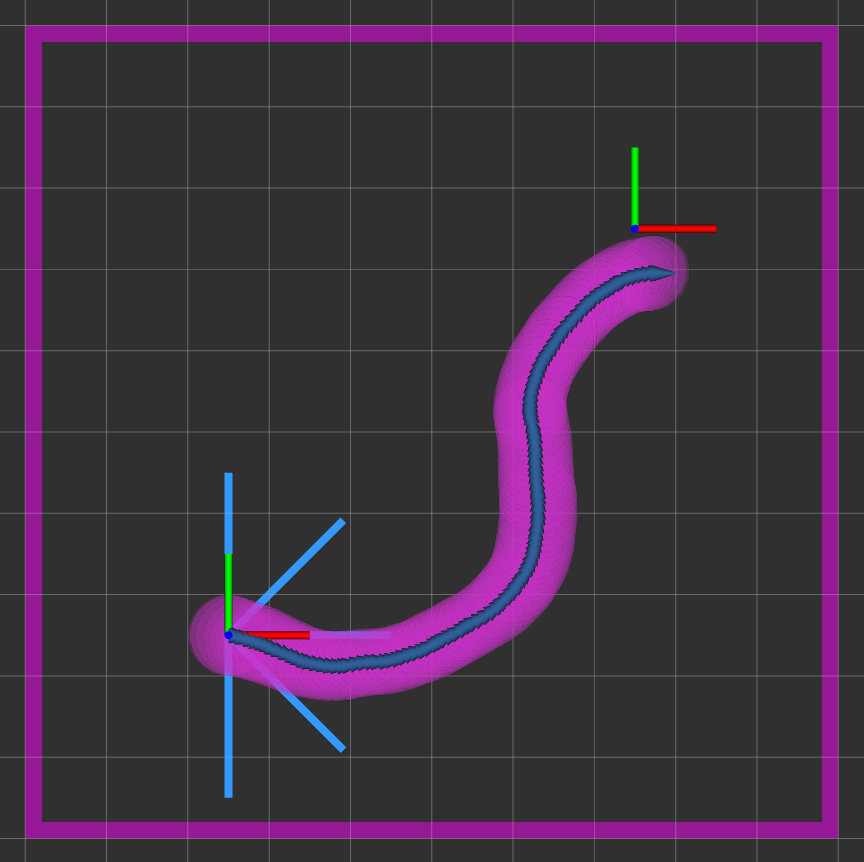}\label{subfig: boundary_initial_traj}}
  \subfloat[]{\includegraphics[width=0.25\columnwidth]{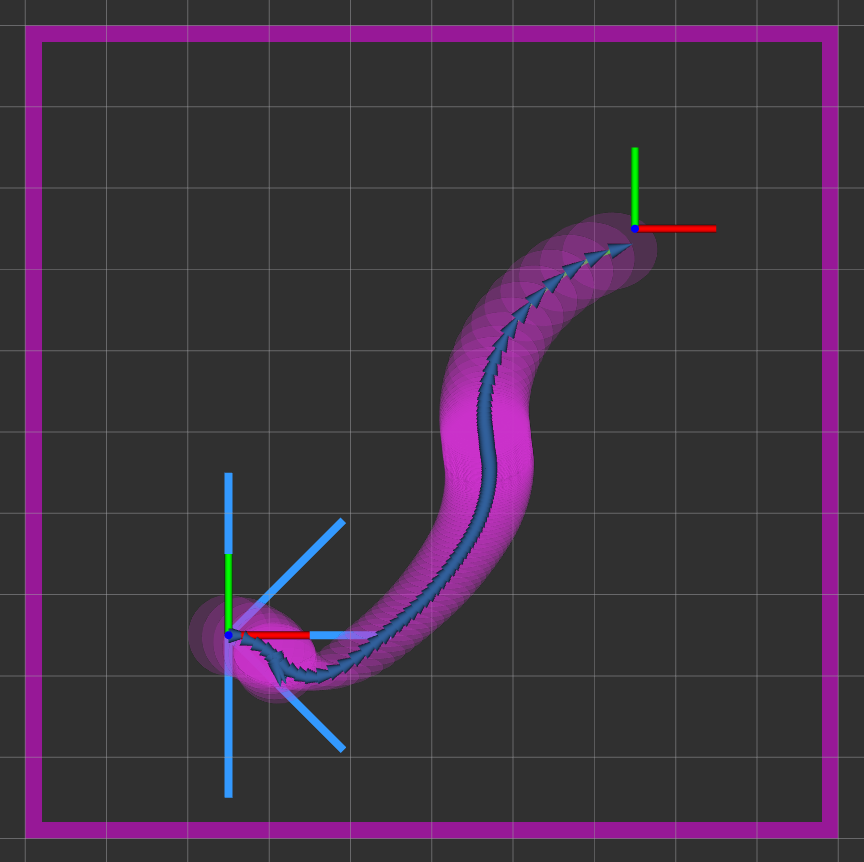}\label{subfig: boundary_nilqg_traj}}
  \subfloat[]{\includegraphics[width=0.25\columnwidth]{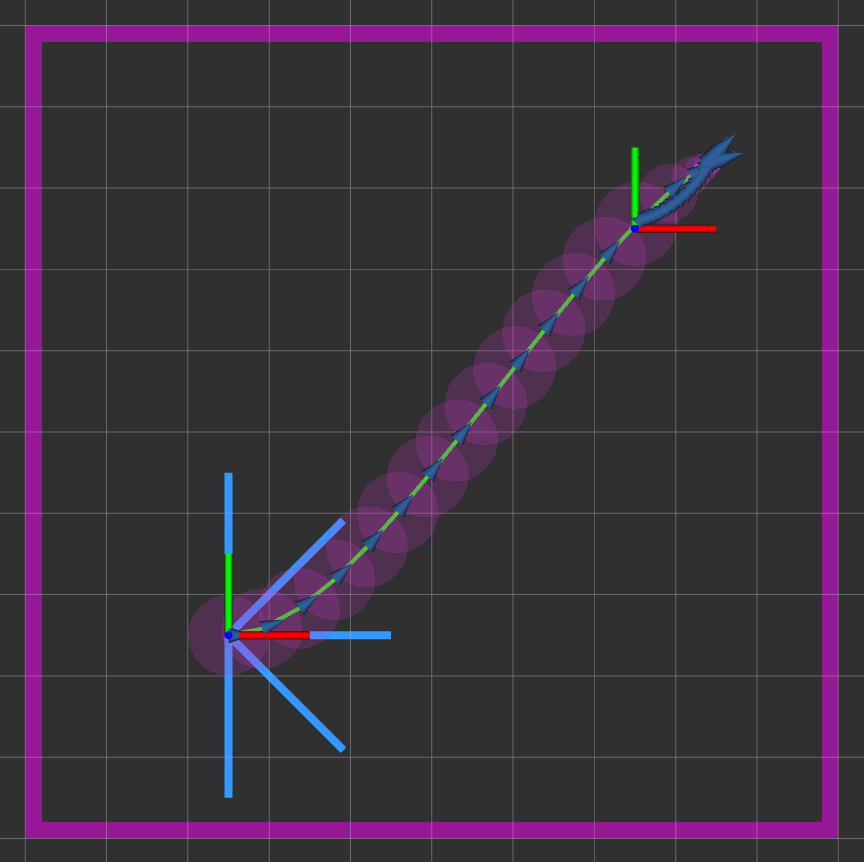}\label{subfig: boundary_eilqg_traj}}
  \subfloat[]{\includegraphics[width=0.25\columnwidth]{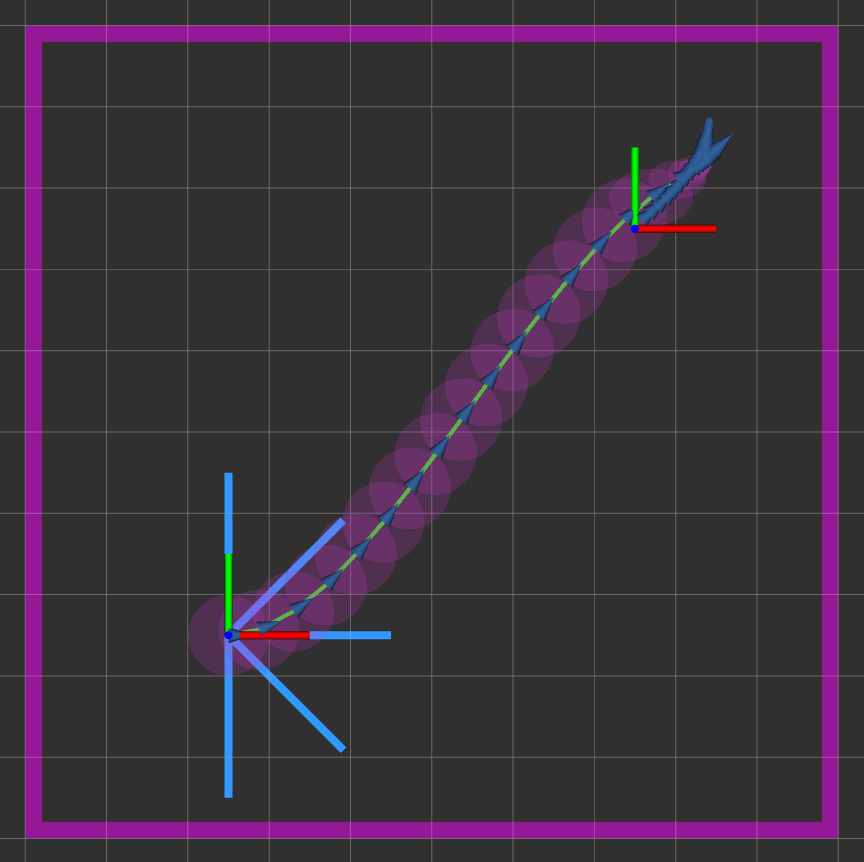}\label{subfig: boundary_uilqg_traj}}
  \caption{Belief nominal trajectories of (a) the initial guess, (b) N-iLQG, (c) E-iLQG, and (d) U-iLQG on a boundary map ($100\times 100$). In each figure, the frames at the bottom left and upper right mark the start and goal locations. The \textcolor{Cyan}{light blue lines} represent five range sensor beams at the maximum range ($2$m). The \textcolor{NavyBlue}{dark blue arrows} and \textcolor{Purple}{purple ellipses} mark the mean and one-standard-deviation position uncertainty.}
  \label{fig: boundary map simulations}
\end{figure}

\begin{table}[t]
  \centering
  \begin{threeparttable}[t]
    \caption{Comparing {\upshape N-iLQG}, {\upshape E-iLQG}, and {\upshape U-iLQG} on the boundary map\tnote{1}}
    \label{table: boundary map simulations}
    \renewcommand{\arraystretch}{1.3}
    \begin{tabular}{c|c|c|c|c|c|c}
      \hline\hline
      method
      & $\mu_0$/$\nu_0$
      & $\lambda_{\mu}$/$\lambda_{\nu}$
      & $\mathbb{E}(c)$
      & CR\tnote{2}
      & iter.\tnote{3}
      & time (s) \\
      \hline
                      N-iLQG  &          $1e3/1e3$\tnote{4} &                        $-$ & $2229.89$ & $0.00$ & $200$\tnote{5} & $14.71$ \\ \hline
                      E-iLQG  &                  $10.0/5.0$ &                  $2.0/2.0$ &  $872.94$ & $0.02$ & $112$          &  $9.13$ \\ \hline
      \multirow{5}{*}{U-iLQG} & \multirow{3}{*}{$10.0/5.0$} &                  $1.2/1.2$ &  $853.04$ & $0.02$ & $227$          & $47.90$ \\ \cline{3-7}
                              &                             &                  $2.0/2.0$ &  $850.78$ & $0.01$ &  $91$          & $16.79$ \\ \cline{3-7}
                              &                             &                  $4.0/4.0$ &  $861.97$ & $0.02$ &  $82$          & $13.61$ \\ \cline{2-7}
                              &                   $5.0/2.5$ & \multirow{2}{*}{$2.0/2.0$} & $1581.43$ & $0.00$ & $143$          & $21.87$ \\ \cline{2-2}\cline{4-7}
                              &                 $20.0/10.0$ &                            & $1802.00$ & $0.00$ & $157$          & $22.05$ \\ \cline{2-2}\cline{4-7}

      \hline\hline
    \end{tabular}
    \begin{tablenotes}
      \item [1] The cost and collision rate are evaluated over $100$ Monte Carlo simulations.
      \item [2] CR refers to collision rate.
      \item [3] The total number of iLQG iterations summed over all outer loops if any.
                The inner loops of iLQG that adjust the damping factor of the Levenberg–Marquardt algorithm are not counted.
      \item [4] For N-iLQG, we directly set $\mu_0=\mu_m=1e3$ and $\nu_0=\nu_m=1e3$.
      \item [5] For each iLQG optimization, the number of iterations is capped at $200$, which is otherwise terminated based on absolute or relative cost reduction.
    \end{tablenotes}
  \end{threeparttable}
\end{table}

We perform an ablation study to demonstrate the effectiveness of the proposed modifications.
In the first variant, we naively apply iLQG on the measurement model~\eqref{eq: range sensor model}.
In the second variant, we use an EKF to approximate the belief dynamics, where numerical differentiation is applied to approximate the measurement Jacobians.
In the following, we refer to the two variants as N-iLQG and E-iLQG.
The proposed method with both modifications is named as U-iLQG.

As show in \figurename~\ref{fig: boundary map simulations}, N-iLQG is not able to utilize the boundaries for localization without the sigmoid function approximation.
In contrast, both E-iLQG and U-iLQG are able to utilize the top right corner to reduce the localization uncertainty, even though the structure is not within the measurement range of the initial trajectory.
Comparing E-iLQG and U-iLQG, the nominal trajectories in the optimized feedback control policies are similar.
However, the lower cost of U-iLQG, shown in Table~\ref{table: boundary map simulations} (comparing the entries with $\mu_0=10.0$, $\nu_0=5.0$, and $\lambda_{\mu}=\lambda_{\nu}=2.0$), confirms that UKFs model the belief dynamics more accurately.

Meanwhile, we study the effect of the parameters in the sigmoid function on U-iLQG.
In this work, a reasonable combination of the parameters, $\mu_0=10.0$, $\nu_0=5.0$, $\lambda_{\mu}=\lambda_{\nu}=2.0$, is determined through trial and error.
By changing the parameters in the neighborhood, it could be observed in Table~\ref{table: boundary map simulations} that the optimized cost is more sensitive to the initial values, $\mu_0$ and $\nu_0$, compared to the scaling factors, $\lambda_{\mu}$ and $\lambda_{\nu}$.
Determining the parameters systematically and scheduling $\mu$ and $\nu$ adaptively could be promising future research directions.

\subsection{Real World Environments}
\label{subsec: real world environments}
\begin{figure*}
  \centering
  \subfloat[]{\includegraphics[width=0.5\textwidth]{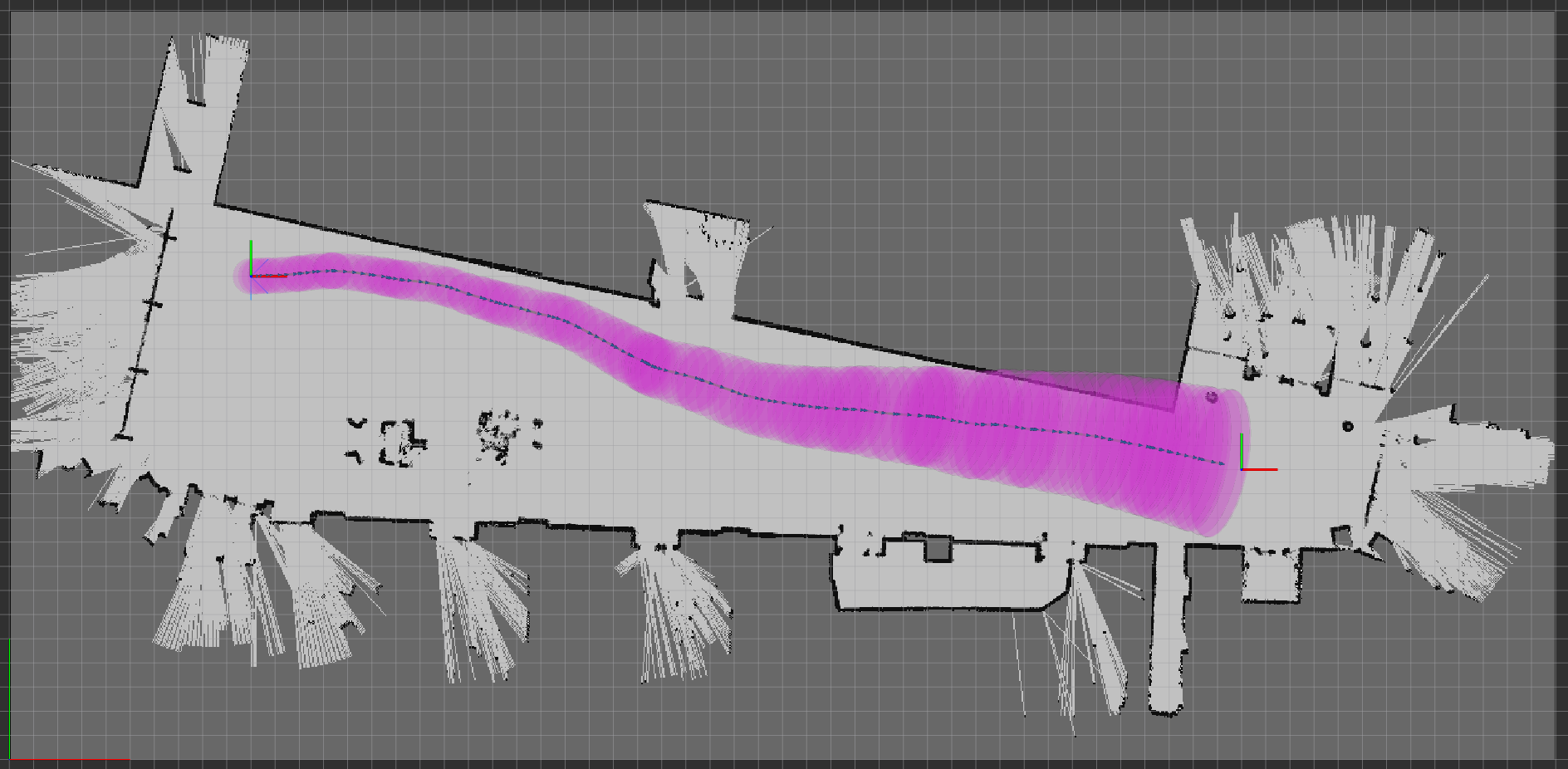}\label{subfig: fr101_initial_traj}}
  \subfloat[]{\includegraphics[width=0.5\textwidth]{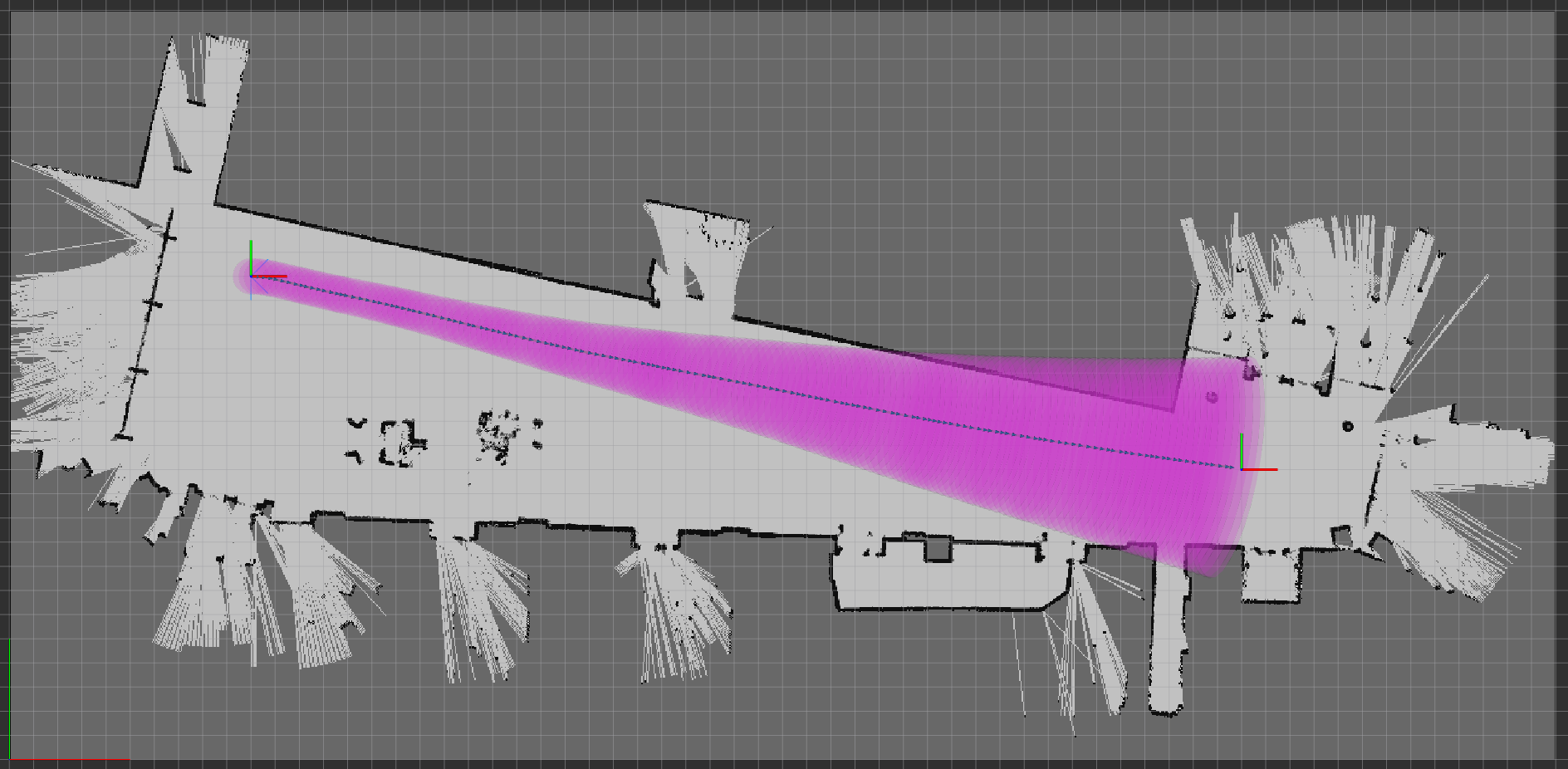}\label{subfig: fr101_ilqr_traj}} \\
  \subfloat[]{\includegraphics[width=0.5\textwidth]{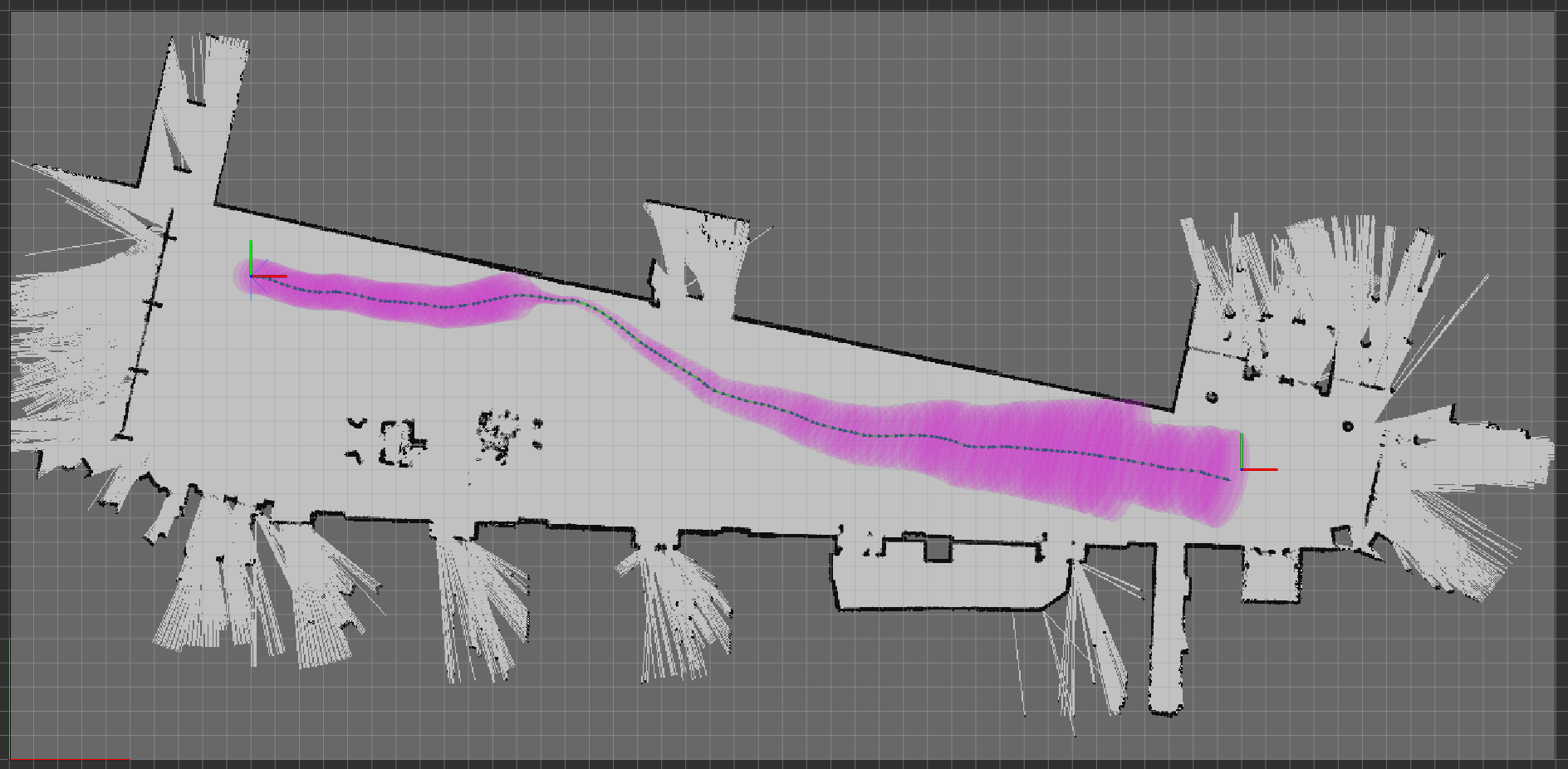}\label{subfig: fr101_milqg_traj}}
  \subfloat[]{\includegraphics[width=0.5\textwidth]{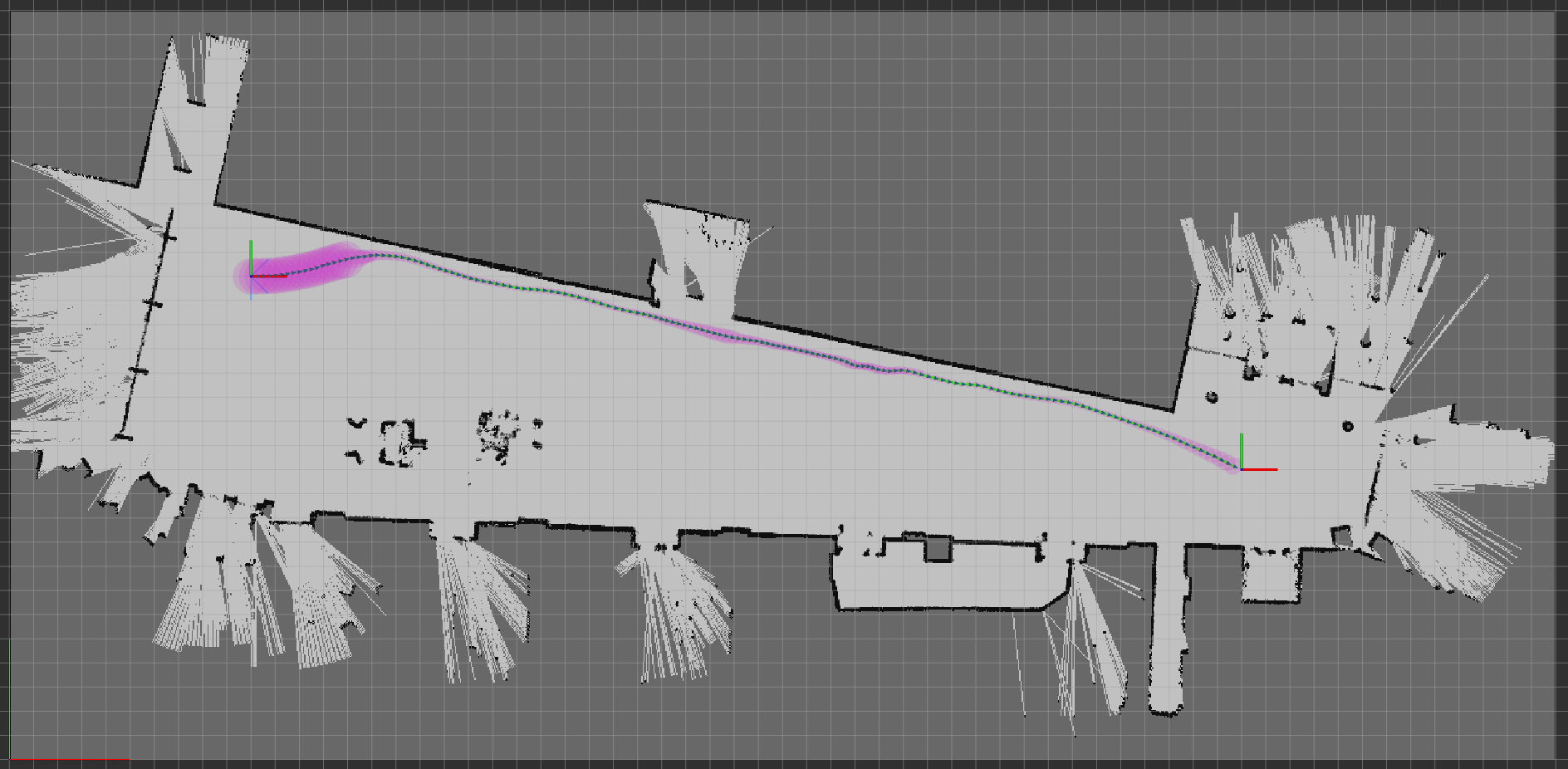}\label{subfig: fr101_uilqg_traj}} \\
  \subfloat[]{\includegraphics[width=0.25\textwidth]{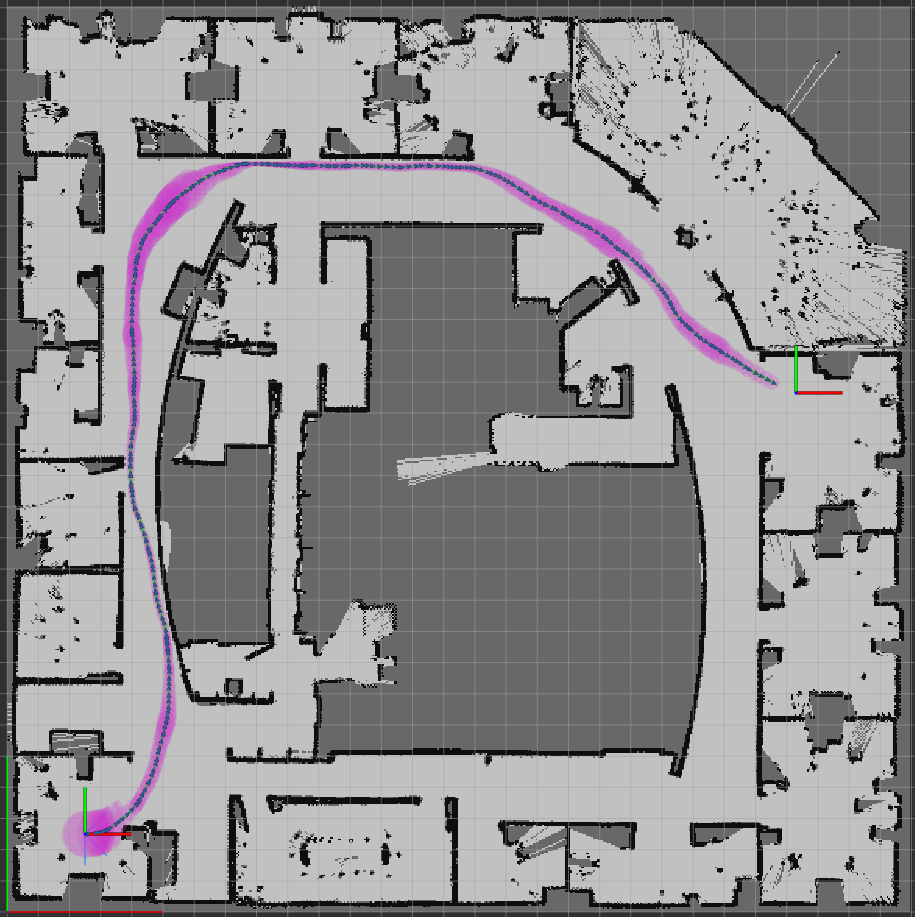}\label{subfig: intel_initial_traj}}
  \subfloat[]{\includegraphics[width=0.25\textwidth]{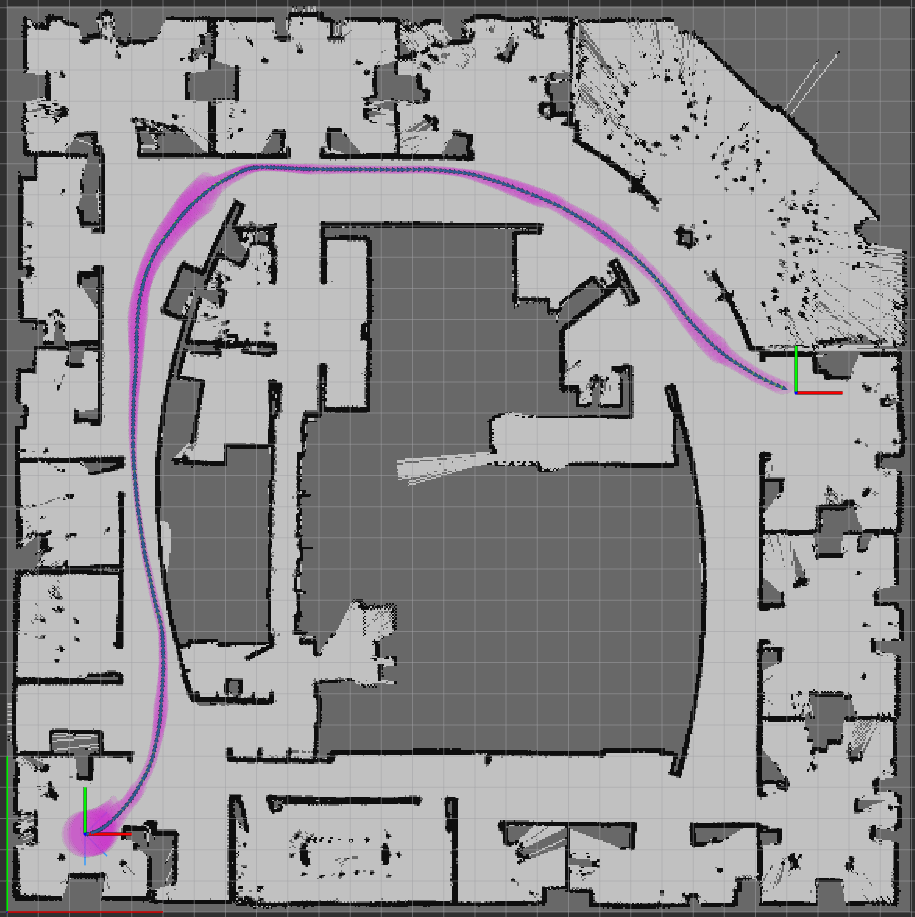}\label{subfig: intel_ilqr_traj}}
  \subfloat[]{\includegraphics[width=0.25\textwidth]{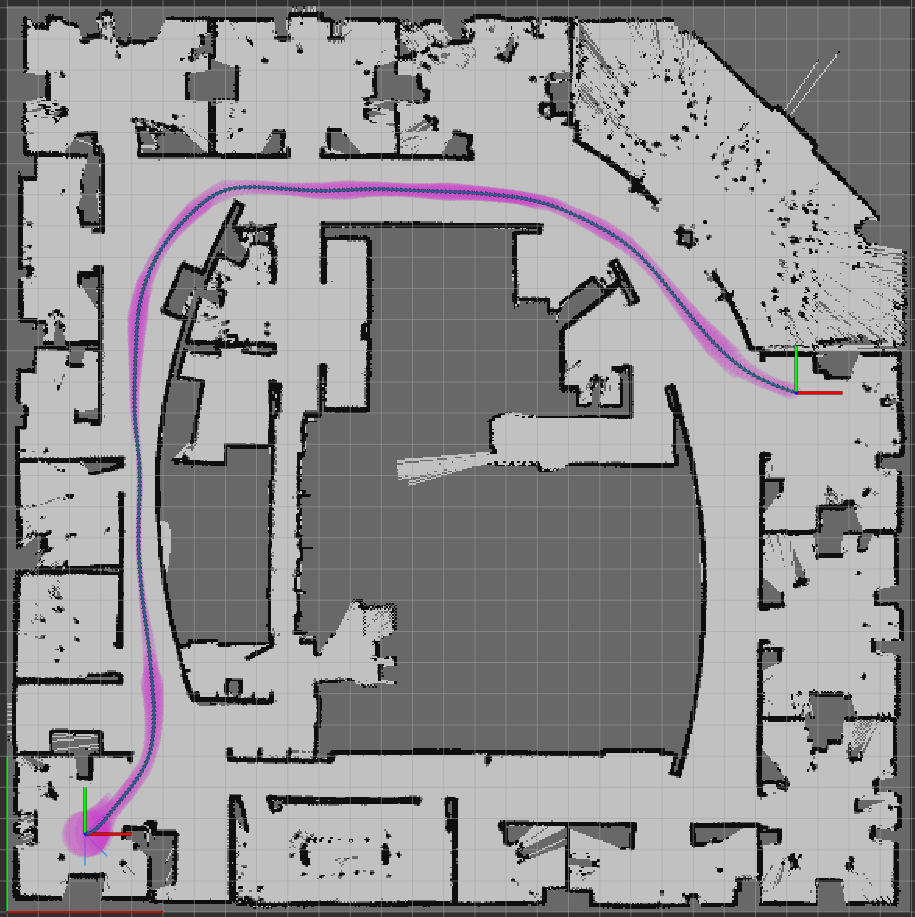}\label{subfig: intel_milqg_traj}}
  \subfloat[]{\includegraphics[width=0.25\textwidth]{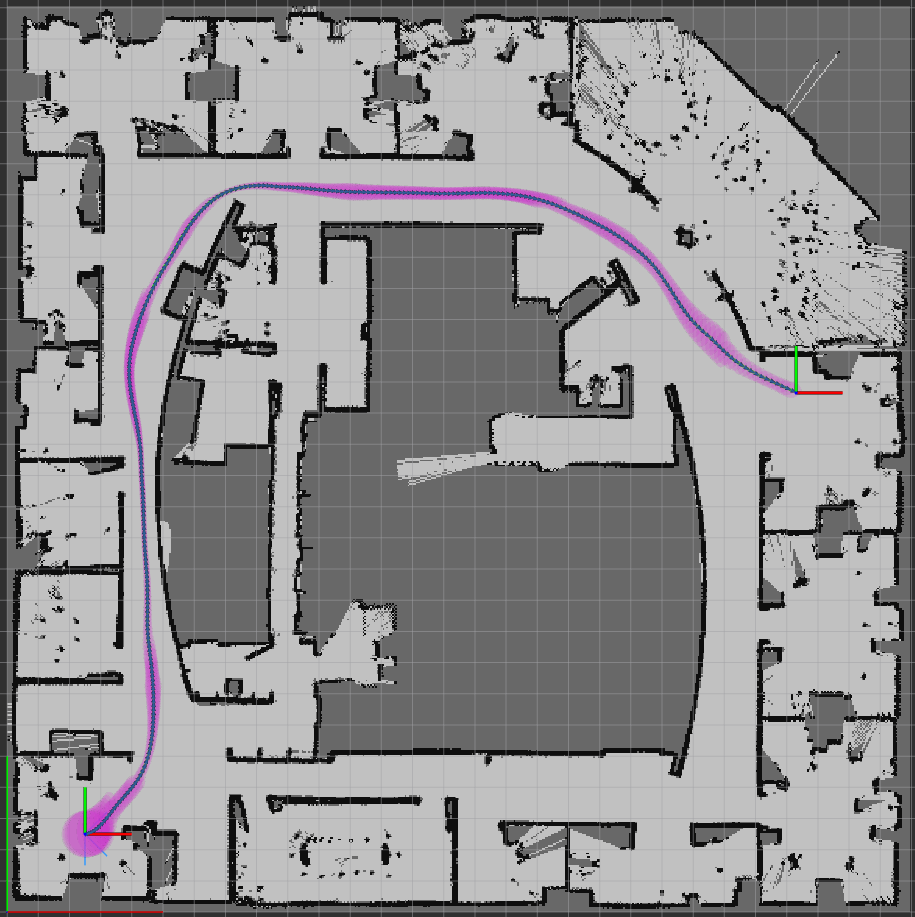}\label{subfig: intel_uilqg_traj}} \\
  \caption{Belief nominal trajectories of (a, e) the initial guess, (b, f) iLQG, (c, g) M-iLQG, and (d, h) U-iLQG in real world environments (a-d) ``fr101'' ($1279\times 620$) and (e-h) ``intel'' ($579\times 581$). In each figure, the frames on the left and right mark the start and goal location. The \textcolor{NavyBlue}{dark blue arrows} and \textcolor{Purple}{purple ellipses} mark the mean and three-standard-deviation position uncertainty.}
  \label{fig: real world environment simulations}
\end{figure*}

\begin{table}[t]
  \centering
  \begin{threeparttable}[t]
    \caption{Comparing {\upshape iLQR}, {\upshape M-iLQG}, and {\upshape U-iLQG} on map {\upshape ``fr101''} and {\upshape ``intel''}\tnote{1, 2}}
    \label{table: real world simulations}
    \renewcommand{\arraystretch}{1.3}
    \begin{tabular}{c|c|c|c|c|c|c}
      \hline\hline
      map
      & method
      & $\mu_0$/$\nu_0$
      & $\mathbb{E}(c)$
      & CR
      & iter.
      & time (s) \\ \hline
      \multirow{3}{*}{fr101} &   iLQR &           $-$ & $1.90e4$ & $0.00$ & $15$ &  $58.94$ \\ \cline{2-7}
                               & M-iLQG &   $50.0/25.0$ & $9.17e3$ & $0.00$ & $34$ & $654.56$ \\ \cline{2-7}
                               & U-iLQG &   $50.0/25.0$ & $3.26e3$ & $0.00$ & $26$ & $703.41$ \\ \hline
      \multirow{3}{*}{intel} &   iLQR &           $-$ & $2.15e3$ & $0.06$ & $11$ &  $45.72$ \\ \cline{2-7}
                               & M-iLQG & $250.0/125.0$ & $1.92e3$ & $0.22$ & $25$ & $369.03$ \\ \cline{2-7}
                               & U-iLQG & $250.0/125.0$ & $1.79e3$ & $0.00$ & $22$ & $631.78$ \\ \hline
      \hline
    \end{tabular}
    \begin{tablenotes}
      \item [1] See Table~\ref{table: boundary map simulations} for related notes.
      \item [2] For M-iLQG and U-iLQG, all simulations use $\lambda_{\mu} = \lambda_{\nu} = 2.0$.
    \end{tablenotes}
  \end{threeparttable}
\end{table}

We also apply the proposed method to large scale maps of real world environments constructed with 2-D Lidar~\cite{Stachniss-Dataset-2020}.
The two maps, named ``fr101'' ($1279\times 620$) and ``intel'' ($579\times 581$), are representative for environments of different clutteredness.
In order to be used in this work, the probabilistic cells in the map are classified as free, occupied, or unknown through thresholding.
The unknown cells are treated as occupied for collision detection in the motion model, while treated as free for ray casting in the measurement model.

The performance of the proposed method is compared with existing state-of-the-art methods that are often applied in practice.
As reviewed in Sec.~\ref{sec: related works}, the separation principle is often extended to nonlinear systems to solve stochastic motion planning problems.
In this case, iLQR~\cite{Li-ICINCO-2004} is directly applied to~\eqref{eq: car-like mobile robot model}.
At the online phase, a UKF is used to estimate belief, the mean of which is fed back to the iLQR policy to generate control inputs.
In the following, we refer to this method as iLQR.
We also compare the proposed method with methods that assume maximum likelihood measurements.
The assumption is equivalent to assuming $\bm{w}_{t+1}=\bm{0}$ in~\eqref{eq: UKF belief dynamics}.
We refer to this method as M-iLQG in the following.
Note that the modifications in Sec.~\ref{subsec: UKF belief dynamics} and~\ref{subsec: densify informative measurements} are also applied to M-iLQG in the comparison.

\begin{figure}[t]
  \centering
  \subfloat[]{\includegraphics[width=0.5\columnwidth]{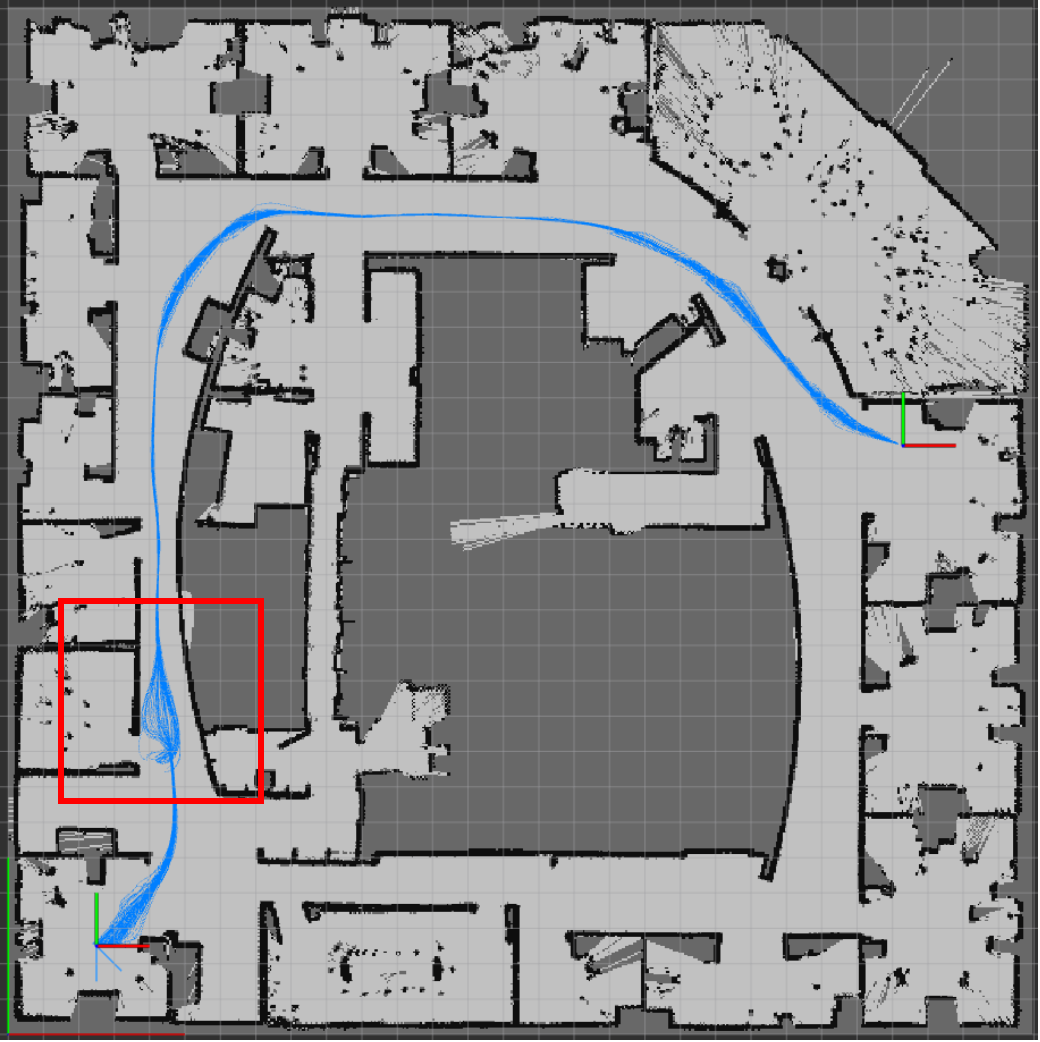} \label{subfig: intel_milqg_sim}}
  \subfloat[]{\includegraphics[width=0.5\columnwidth]{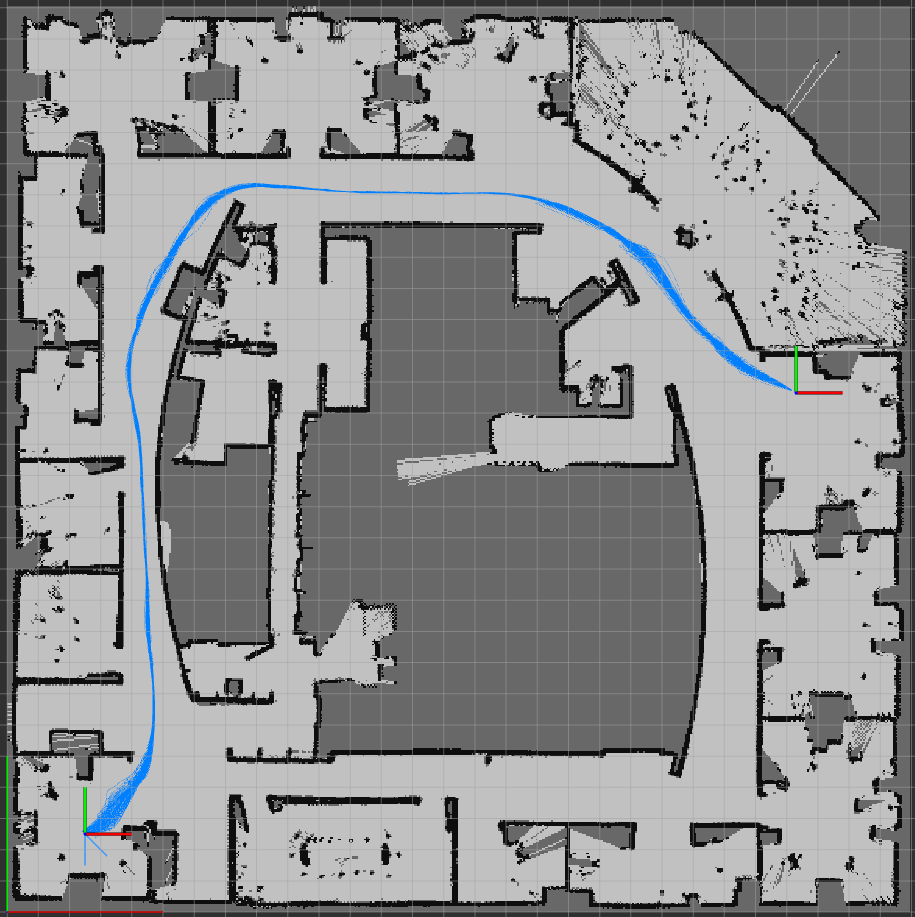}\label{subfig: intel_uilqg_sim}}
  \caption{Mean of simulated belief trajectories using policies optimized with (a) M-iLQG and (b) U-iLQG on ``intel''. Note \textcolor{red}{the marked region} in (a). The robot just enters the corridor with relatively large uncertainty. The measurements collected at the doorway induce large innovation, which leads to instability of the policy optimized with M-iLQG. The same problem does not appear with U-iLQG.}
  \label{fig: intel simulated beliefs}
\end{figure}

As shown in \figurename~\ref{fig: real world environment simulations}a-d, the feedback policy of U-iLQG is able to actively localize the robot by moving along the wall in an open environment.
Therefore, as shown in Table~\ref{table: real world simulations}, the cost of U-iLQG is significantly lower compared with the other two methods.
In a more cluttered environment, \figurename~\ref{fig: real world environment simulations}e-h, active localization is no longer necessary.
The nominal trajectories are similar for all methods.
However, Table~\ref{table: real world simulations} shows the collision rate of iLQR and M-iLQG are higher compared to U-iLQG.
The high collision rate has different causes.
In iLQR, the nominal trajectory is over close to the wall since the state estimation uncertainty is ignored.
For M-iLQG, the feedback policy is less robust to innovation noise as a result of assuming deterministic belief dynamics.
Especially, when $\partial \bm{b}_{t+1} / \partial \bm{w}_{t+1}$ is large, the nonzero innovation may result in a large update in belief which deviates significantly from the nominal trajectory, and possibly leads to collisions in M-iLQG.
\figurename~\ref{fig: intel simulated beliefs} illustrates the issue of M-iLQG with simulated belief trajectories.

\section{Conclusion}
\label{sec: conclusion}
In this work, we apply iLQG to generate local optimal control policies to navigate a car-like robot with low-cost range sensors.
We introduce two modifications addressing issues which prevent applications of iLQG with the considered robotic system.
First, we use a UKF to model the belief dynamics.
As a result, the requirements for differentiable motion and measurement models are removed.
Second, we use a sigmoid function to smooth the discontinuity of the noise standard deviation in the range sensor model.
By iteratively updating the parameters of the sigmoid function in the optimization, the approximated range sensor model eventually converges to the true model.
We demonstrate the effectiveness of the modifications through an ablation study.
We also show superior performance of the proposed method through comparisons with the state-of-the-art methods in real world environments.

The proposed method is subject to a few limitations which could potentially be addressed in the future work.
As noted in Sec.~\ref{subsec: ablation study}, the solution of iLQG is sensitive to the initial values of the parameters in the sigmoid function.
In the future work, it is promising to investigate a systematic way to select and adaptively update of the sigmoid function parameters to ensure reliable convergence.
Another limitation is that the task has to be completely modeled by an objective function.
Not only variables like controls are subject to hard constraints, but the different physical units of costs make the cost parameters hard to tune.
The generality of iLQG framework can be improved if constraints can be introduced into the optimization.

\appendix[Differentiability of the belief dynamics]
The true discrete-time belief dynamics is modeled by the Bayes filter in the following form,
\begin{equation}
  \label{eq: bayes filter}
  \begin{aligned}
    &\bm{p}_{t+1}(\bm{x}) = \int t(\bm{x}|\bm{y}, \bm{u}_t) \bm{b}_t(\bm{y}) d\bm{y}, \\
    &\bm{q}_{t+1}(\bm{x}) = l(\bm{z}_{t+1}|\bm{x}) \bm{p}_{t+1}(\bm{x}),\\
    &\bm{b}_{t+1}(\bm{x}) = \frac{1}{\int \bm{q}_{t+1}(\bm{y}) d\bm{y}} \bm{q}_{t+1}(\bm{x}).
  \end{aligned}
\end{equation}
In~\eqref{eq: bayes filter}, $t(\cdot)$ is the transition probability.
$l(\cdot)$ is the measurement likelihood.
$\bm{b}_t$ is the belief at $t$.
$\bm{p}_{t+1}$ is the prior density. $\bm{q}_{t+1}$ is the unnormalized posterior density, normalizing which gives the belief at $t+1$, $\bm{b}_{t+1}$.

\begin{assumption}
  \label{assumption: assumptions of the system}
  Define $\mathcal{X}$, $\mathcal{U}$, and $\mathcal{Z}$ as the state, control, and measurement spaces respectively. We assume the following,
  \renewcommand{\labelenumi}{(\arabic{enumi})}
  \begin{enumerate}
    \item $t(\cdot)$ is continuous on $\mathcal{X}^2 \times \mathcal{U}$, and $0 < t(\cdot) < \infty$.
    \item $\partial t/\partial \bm{u}_t$ exists and is continuous on $\mathcal{X}^2 \times \mathcal{U}$.
    \item $l(\cdot)$ is continuous on $\mathcal{X} \times \mathcal{Z}$, and $0 < l(\cdot) < \infty$.
    \item $\partial l/ \partial \bm{z}_{t+1}$ exists.
    \item $\bm{b}_{t}$ is continuous on $\mathcal{X}$.
  \end{enumerate}
\end{assumption}

The above are weak assumptions of a system.
The first four items are trivially satisfied by assuming the motion and measurement noises are Gaussian, as in~\eqref{eq: general discrete time system motion and measurement models}.
Meanwhile, we show in the following that the continuity of $\bm{b}_t$ can be preserved by~\eqref{eq: bayes filter}.
Therefore, it is sufficient to assume the continuity of $\bm{b}_0$ in order to ensure the continuity of $\bm{b}_t$'s for future steps.

\begin{theorem}[Continuity of belief]
  \label{theorem: continuity of belief}
  $\bm{b}_{t+1}$ is continuous on $\mathcal{X}$.
\end{theorem}

\begin{IEEEproof}
  Since $t(\cdot)$, $l(\cdot)$, and $\bm{b}_t$ are continuous functions, $\bm{p}_{t+1}$ and $\bm{q}_{t+1}$ are continuous.
  Given that $0 < t(\cdot),\ l(\cdot) < \infty$ and $\int \bm{b}_t(\bm{y}) d\bm{y} = 1$, we have $0 < \int \bm{q}_{t+1}(\bm{y}) d\bm{y} < \infty$.
  Therefore, $\bm{b}_{t+1}$ is continuous.
\end{IEEEproof}

Next we show the differentiability of $\bm{b}_{t+1}$ \textit{w.r.t.} $\bm{u}_t$, $\bm{z}_t$, and $\bm{b}_{t}$.
Note that $\bm{b}_{t+1}$ is a probability density function.
It does not make sense to discuss the differentiability of a function \textit{w.r.t.} vectors or other functions directly.
Instead, we consider the pointwise differentiability, \textit{i.e.}, the relationships between $d\bm{b}_{t+1}(\bm{x})$ and $d\bm{u}_t$, $d\bm{z}_{t+1}$, and $d\bm{b}_t$.
Specially, $\bm{b}_{t+1}(\bm{x})$ should be considered as a functional of $\bm{b}_t$.

\begin{theorem}[Differentiability of the discrete-time belief dynamics]
  \label{theorem: differentiability of discrete-time belief dynamics}
  $\bm{b}_{t+1}$ is pointwise differentiable \textit{w.r.t.} $\bm{u}_t$, $\bm{z}_t$, and $\bm{b}_{t}$.
\end{theorem}

\begin{IEEEproof}
  The differentiability of $\bm{b}_{t+1}(\bm{x})$ could be shown by construction.
  \small
  \begin{align*}
    d\bm{p}_{t+1}(\bm{x}) &= \int \frac{\partial t}{ \partial \bm{u}_t} (\bm{x}|\bm{y}, \bm{u}_t) d\bm{u}_t \bm{b}_t(\bm{y}) d\bm{y}
                          + \int t(\bm{x}|\bm{y}, \bm{u}_t) d\bm{b}_t(\bm{y}) d\bm{y}, \\
    d\bm{q}_{t+1}(\bm{x}) &= \frac{\partial l}{\partial \bm{z}_{t+1}} (\bm{z}_{t+1}|\bm{x}) d\bm{z}_{t+1} \bm{p}_{t+1}(\bm{x})
                          + l(\bm{z}_{t+1}|\bm{x}) d\bm{p}_{t+1}(\bm{x}), \\
    d\bm{b}_{t+1}(\bm{x}) &= \frac{d\bm{q}_{t+1}(\bm{x})}{\int \bm{q}_{t+1}(\bm{y}) d\bm{y}}
                          - \frac{\bm{q}_{t+1}(\bm{x}) \cdot \int d\bm{q}_{t+1}(\bm{y}) d\bm{y}}{\prl{\int \bm{q}_{t+1}(\bm{y}) d\bm{y}}^2}.
  \end{align*}
  \normalsize
  The differential equations hold because of Assumption~\ref{assumption: assumptions of the system}, which allows the application of Leibniz's rule to switch the order of differentiation and integration.
  Applying the chain rule produces the partial derivatives, $\partial \bm{b}_{t+1}(\bm{x}) / \partial \bm{b}_t(\bm{y})$, $\partial \bm{b}_{t+1}(\bm{x}) / \partial \bm{u}_t$, and $\partial \bm{b}_{t+1}(\bm{x}) / \partial \bm{z}_{t+1}$.
\end{IEEEproof}

Therefore, the differentiability of belief dynamics for systems in~\eqref{eq: general discrete time system motion and measurement models} is independent of $f(\cdot)$ and $h(\cdot)$.

In Gaussian filters, such as EKFs and UKFs, (multivariate) Gaussian distributions are used to approximate the underlying belief with the first and second moments.
In the following, we show the differentiability of the first two moments of $\bm{b}_{t+1}$ by construction, assuming $\bm{b}_t$ is Gaussian.

\begin{corollary}[Differentiability of the first two moments]
  \label{corollary: differentiability of first and second moments}
  The first two moments of $\bm{b}_{t+1}$ are differentiable \textit{w.r.t.} $\bm{u}_t$ and $\bm{z}_t$, and first two moments of $\bm{b}_t$ assuming $\bm{b}_t$ is Gaussian.
\end{corollary}

\begin{IEEEproof}
  Consider the mean, $\mathbb{E}\crl{\bm{x}_{t+1}}$ of $\bm{b}_{t+1}$.
  Differentiating $\mathbb{E}\crl{\bm{x}_{t+1}}$ \textit{w.r.t.} $\bm{b}_{t+1}$ while applying Theorem~\ref{theorem: differentiability of discrete-time belief dynamics} gives,
  \begin{equation}
    \label{eq: differentiating first moment}
    \begin{aligned}
      d\mathbb{E}\crl{\bm{x}_{t+1}} &= \int \bm{x} \cdot d\bm{b}_{t+1}(\bm{x}) d\bm{x},\\
      d\bm{b}_{t+1}(\bm{x}) &= \frac{\partial \bm{b}_{t+1}(\bm{x})}{\partial \bm{u}_t} d\bm{u}_t +
                               \frac{\partial \bm{b}_{t+1}(\bm{x})}{\partial \bm{z}_{t+1}} d\bm{z}_{t+1} \\
                            &+ \int \frac{\partial \bm{b}_{t+1}(\bm{x})}{\partial \bm{b}_t(\bm{y})} d\bm{b}_t(\bm{y}) d\bm{y}.
    \end{aligned}
  \end{equation}
  $\bm{b}_t$ is an (multivariate) Gaussian parameterized by $\mathbb{E}\crl{\bm{x}_t}$ and $\mathbb{E}\crl{\bm{x}\bm{x}^\top}$.
  Therefore,
  \begin{equation}
    \label{eq: differentiating multivariate gaussian}
    d\bm{b}_t(\bm{x}) = \frac{\partial \bm{b}_t(\bm{x})}{\partial \mathbb{E}\crl{\bm{x}_t}} d\mathbb{E}\crl{\bm{x}_t} +
                        \frac{\partial \bm{b}_t(\bm{x})}{\partial \mathbb{E}\crl{\bm{x}_t \bm{x}_t^\top}} d\mathbb{E}\crl{\bm{x}_t \bm{x}_t^\top}.
  \end{equation}
  Details of $\partial \bm{b}_t(\bm{x}) / \partial \mathbb{E}\crl{\bm{x}_t}$ and $\partial \bm{b}_t(\bm{x}) / \partial \mathbb{E}\crl{\bm{x}_t \bm{x}_t^\top}$ are provided in~\cite{Petersen-Denmark-2012}.
  Combining~\eqref{eq: differentiating first moment} and~\eqref{eq: differentiating multivariate gaussian} produces $\partial \mathbb{E}\crl{\bm{x}_{t+1}} / \partial \mathbb{E}\crl{\bm{x}_{t}}$ and $\partial \mathbb{E}\crl{\bm{x}_{t+1}} / \partial \mathbb{E}\crl{\bm{x}_t \bm{x}_t^\top}$.
  Similar steps can be applied to obtain $\partial \mathbb{E}\crl{\bm{x}_{t+1} \bm{x}_{t+1}^\top} / \partial \mathbb{E}\crl{\bm{x}_{t}}$ and $\partial \mathbb{E}\crl{\bm{x}_{t+1} \bm{x}_{t+1}^\top} / \partial \mathbb{E}\crl{\bm{x}_t \bm{x}_t^\top}$, which completes the proof.
\end{IEEEproof}

As a result of Corollary~\ref{corollary: differentiability of first and second moments}, if $\bm{b}_{t+1}$ is approximated as a Gaussian distribution using the first two moments, the parametric representation of the belief is differentiable.

\bibliographystyle{IEEEtran}
\bibliography{IEEEabrv,ref}

\end{document}